\pdfoutput=1

\documentclass[11pt]{article}

\usepackage[]{naacl2021}

\usepackage{times}
\usepackage{latexsym}

\usepackage{microtype}
\usepackage{graphicx,epsfig,subfig}
\usepackage{algorithm}
\usepackage{algorithmic}
\usepackage{hyperref}
\usepackage{multirow}
\usepackage{rotating}

\usepackage[T1]{fontenc}

\usepackage[utf8]{inputenc}

\usepackage{microtype}

\title{Unsupervised Self-Training for Sentiment Analysis of Code-Switched Data}

\author{Akshat Gupta$^1$, Sargam Menghani$^1$, Sai Krishna Rallabandi$^2$, Alan W Black$^2$ \\
         $^1$Department of Electrical and Computer Engineering, Carnegie Mellon University\\ $^2$Language Technologies, Institute, Carnegie Mellon University \\ \texttt{ \{akshatgu, smenghan\}@andrew.cmu.edu, \{srallaba, awb\}@cs.cmu.edu}}

\begin{document}
\maketitle

\begin{abstract}
Sentiment analysis is an important task in understanding social media content like customer reviews, Twitter and Facebook feeds etc. In multilingual communities around the world, a large amount of social media text is characterized by the presence of code-switching. Thus, it has become important to build models that can handle code-switched data. However, annotated code-switched data is scarce and there is a need for unsupervised models and algorithms. We propose a general framework called Unsupervised Self-Training and show its applications for the specific use case of sentiment analysis of code-switched data. We use the power of pre-trained BERT models for initialization and fine-tune them in an unsupervised manner, only using pseudo labels produced by zero-shot transfer. We test our algorithm on multiple code-switched languages and provide a detailed analysis of the learning dynamics of the algorithm with the aim of answering the question - `Does our unsupervised model understand the Code-Switched languages or does it just learn its representations?'. Our unsupervised models compete well with their supervised counterparts, with their performance reaching within 1-7\% (weighted F1 scores) when compared to supervised models trained for a two class problem. 
\end{abstract}

\section{Introduction}
Sentiment analysis, sometimes also known as opinion mining, aims to understand and classify the opinion, attitude and emotions of a user based on a text query. Sentiment analysis has many applications including understanding product reviews, social media monitoring, brand monitoring, reputation management etc. Code switching is referred to as the phenomenon of alternation between multiple languages, usually two, within a single utterance. Code switching is very common in many bilingual and multilingual societies around the world including India (Hinglish, Tanglish etc.), Singapore (Chinglish) and various Spanish speaking areas of North America (Spanglish). A large amount of social media text in these regions is code-mixed, which is why it is essential to build systems that are able to handle code switching. 

Various datasets have been released to aid advancements in Sentiment Analysis of code-mixed data. These datasets are usually much smaller and more noisy when compared to their high-resource-language-counterparts and are available for very few languages. Thus, there is a need to come up with both unsupervised and semi-supervised algorithms to deal with code-mixed data. In our work, we present a general framework called Unsupervised Self-Training Algorithm for doing sentiment analysis of code-mixed data in an unsupervised manner. We present results for four code-mixed languages - Hinglish (Hindi-English), Spanglish (Spanish-English), Tanglish (Tamil-English) and Malayalam-English.

In this paper, we propose the Unsupervised Self-Training framework and apply it to the problem of sentiment classification. Our proposed framework performs two tasks simultaneously - firstly, it gives sentiment labels to sentences of a code-mixed dataset in an unsupervised manner, and secondly, it trains a sentiment classification model in a purely unsupervised manner. The framework can be extended to incorporate active learning almost seamlessly.  We present a rigorous analysis of the learning dynamics of our unsupervised model and try to answer the question - \textit{'Does the unsupervised model understand the code-switched languages or does it just recognize its representations?'}. We also show methods for optimizing performance of the Unsupervised Self-Training algorithm.

\section{Related Work}
In this paper, we propose a framework called Unsupervised Self-Training, which is an extension to the semi-supervised machine learning algorithm called Self-Training \cite{zhu2005semi}. Self-training has previously been used in natural language processing for pre-training \cite{du2020self} and for tasks like word sense disambiguation \cite{yarowsky1995unsupervised}. It has been shown to be very effective for natural language processing tasks \cite{selftraining_nlp_fair} and better than pre training in low resource scenarios both theoretically \cite{selftraining_theoretical} and emperically\cite{selftraining_brain}. \citealt{zoph2020rethinking} show that self-training can be a more useful than pre-training in high resourced scenarios for the task of object detection, and a combination of pre-training and self-training can improve performance when only 20\% of the available dataset was used. However, our proposed framework differs from self-training such that we only use the zero-shot predictions made by our initialization model to train our models and never use actual labels.

Sentiment analysis is a popular task in industry as well as within the research community, used in analysing the markets \cite{nanli2012sentiment}, election campaigns \cite{haselmayer2017sentiment} etc. A large amount of social media text in most bilingual communities is code-mixed and many labelled datasets have been released to perform sentiment analysis. We will be working with four code-mixed datasets for sentiment analysis, Malayalam-English and Tamil-English \citep{chakravarthi2020sentiment, chakravarthi2020corpus, chakravarthi2020overview} and Spanglish and Hinglish \cite{patwa2020semeval}. 

Previous work has shown BERT based models to achieve state of the art performance for code-switched languages in tasks like offensive language identification \cite{jayanthi2021sj_aj} and sentiment analysis \cite{gupta2021task}. We will build unsupervised models on top of BERT. BERT \cite{devlin2018bert} based models have achieved state of the art performance in many downstream tasks due to their superior contextualized representations of language, providing true bidirectional context to word embeddings. We will use the sentiment analysis model from \cite{barbieri2020tweeteval}, trained on a large corpus of English Tweets (60 million Tweets) for initializing our algorithm. We will refer to the sentiment analysis model from \cite{barbieri2020tweeteval} as the \textit{TweetEval} model in the remainder of the paper. The TweetEval model is built on top of an English RoBERTa \cite{liu2019roberta} model.

\begin{figure*}[ht]
  \centering
    \includegraphics[width=\linewidth]{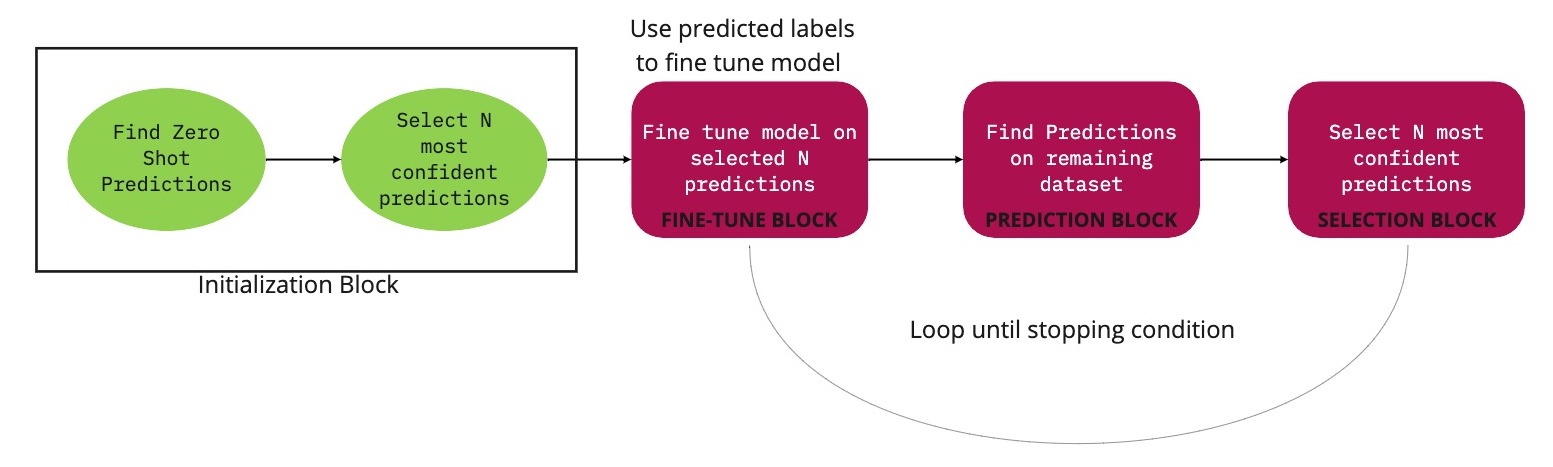}
    \caption{A visual representation of our proposed Unsupervised Self-Training framework.}\label{fig:rec}
\end{figure*}

\section{Proposed Approach: Unsupervised Self-Training}
Our proposed algorithm is centred around the idea of creating an unsupervised learning algorithm that is able to harness the power of cross-lingual transfer in the most efficient way possible, with the aim of producing unsupervised sentiment labels. In its most fundamental form, our proposed Unsupervised Self-Training algorithm\footnote{ The code for the framework can be found here: \href{https://github.com/akshat57/Unsupervised-Self-Training}{https://github.com/akshat57/Unsupervised-Self-Training}} is shown in Figure \ref{fig:rec} is shown in Figure \ref{fig:rec}.

We begin by producing zero-shot results for sentiment classification using a selected pre-trained model trained for the same task. From the predictions made, we select the top-N most confident predictions made by the model. The confidence level is judged by the softmax scores. Making the zero-shot predictions and selecting sentences make up the \textit{Initialization block} as shown in Figure \ref{fig:rec}. We then use the pseduo-labels predicted by the zero-shot model to fine tune our model. After that, predictions are made on the remaining dataset with the fine-tuned model. We again select sentences based on their softmax scores for fine-tuning the model in the next iteration. These steps are repeated until we've gone through the entire dataset or until a stopping condition. At all fine-tuning steps, we only use the predicted pseduo-labels as ground truth to train the model, which makes the algorithm completely unsupervised.

As the first set of predicted pseudo-labels are produced by a zero-shot model, our framework is very sensitive to initialization. Care must be taken to initialize the algorithm with a \textit{compatible model}. For example, for the task of sentiment classification of Hinglish  Twitter data, an example of a compatible initial model would be a sentiment classification model trained on either English or Hindi sentiment data. It would be even more compatible if the model was trained on Twitter sentiment data, the data thus being from the same domain. 



\subsection{Optimizing Performance}
The most important blocks in the Unsupervised Self-Training framework with respect to maximizing performance are the \textit{Initialization Block} and the \textit{Selection Block} (Figure \ref{fig:rec}). To improve initialization, we must choose the most compatible model for the chosen task. Additionally, to improve performance, we can use several training strategies in the Selection Block. In this section we discuss several variants of the Selection Block.

As an example, instead of selecting a fixed number of samples N from the dataset in the selection block,  we could be selecting a different but fixed number $N_i$ from each class $i$ in the dataset. This would need an understanding of the class distribution of the dataset. We discuss this in later sections. Another variable number of sentences in each iteration, rather than a fixed number. This would give us a \textit{selection schedule} for the algorithm. We explore some of these techniques in later sections.

Other factors can be incorporated in the Selection Block. Selection need not be based on \textit{just} the most confident predictions. We can have additional selection criteria, for example, incorporating the Token Ratio (defined in section \ref{section:analysis}) of a particular language in the predicted sentences. Taking the example of a Hinglish dataset, one way to do this would be to select sentences that have a larger amount of Hindi and are within selection threshold. In our experiments, we find that knowing an optimal selection strategy is vital to achieving the maximum performance.


\begin{table*}
\centering
\begin{tabular}{cccccc}
\hline
\multicolumn{1}{|p{1cm}|}{\centering \textbf{Language}} & \multicolumn{1}{|p{1cm}|}{\centering \textbf{Domain}} & 
\multicolumn{1}{|p{1.5cm}|}{\centering \textbf{Total}} & \multicolumn{1}{|p{1.5cm}|}{\centering \textbf{Positive}} & \multicolumn{1}{|p{1.5cm}|}{\centering \textbf{Negative}} & \multicolumn{1}{|p{1.5cm}|}{\centering \textbf{Neutral}}\\
\hline
Hinglish & Tweets & 14000 & 4634 & 4102 & 5264 \\
Spanglish & Tweets & 12002 & 6005 & 2023 & 3974 \\
Tanglish & Youtube Comments & 9684 & 7627 & 1448 & 609\\
Malayalam-English & Youtube Comments & 3915 & 2022 & 549 & 1344\\
\hline
\end{tabular}
\caption{
Training dataset statistics for chosen datasets.
}\label{Table1}
\end{table*}

\section{Datasets} \label{sec:datasets}
We test our proposed framework on four different languages - Hinglish \cite{patwa2020semeval}, Spanglish \cite{patwa2020semeval}, Tanglish \cite{chakravarthi2020corpus} and Malayalam-English \cite{chakravarthi2020sentiment}. The statistics of the training sets are given in Table \ref{Table1}. We also use the test sets of the above datasets, which have similar distribution as their respective training sets. The statistics of the test sets are not shown for brevity. We ask the reader to refer to the respective papers for more details. 

The choice of datasets, apart from covering three language families, incorporate several other important features. We can see from Table \ref{Table1} that the four datasets have different sizes, the Malayalam-English dataset being the smallest. Apart from the Hinglish dataset, the other three datasets are highly imbalanced. This is an important distinction as we cannot expect an unknown set of sentences to have a balanced class distributions. We will later see that having a biased underlying distribution affects the performance of our algorithm and how better training strategies can alleviate this problem.

The chosen datasets are also from two different domains - the Hinglish and Spanglish datasets are a collection of Tweets whereas Tanglish and Malaylam-English are a collection of Youtube Comments. The TweetEval model, which is used for initialization is trained on a corpus of English Tweets. Thus the Hinglish and Spangish datasets are in-domain datasets for our initialization model \cite{barbieri2020tweeteval}, whereas the Dravidian language (Tamil and Malayalam) datasets are out of domain.

The datasets also differ in the amount of class-wise code-mixing. Figure \ref{fig:1} shows that for the Hinglish dataset, a negative Tweet is more likely to contain large amounts of Hindi. This is not the same for the other datasets. For Spanglish, both positive and negative sentiment Tweets have a tendency to use a larger amount of Spanish than English.

An important thing to note here is that each of the four code-mixed datasets selected are written in the latin script. Thus our choice of datasets does not take into account mixing of different scripts.

\section{Models}
Models built on top of BERT \cite{devlin2018bert} and its multilingual version like mBERT, XLM-RoBERTa \cite{conneau2019unsupervised} have recently produced state-of-the-art results in many natural language processing tasks. Various shared tasks \cite{patwa2020semeval} \cite{chakravarthi2020overview} in the domain of code-switched sentiment analysis have also seen their best performing systems build on top of these BERT models. 

English is a common language among all the four code-mixed datasets being considered. This is why we use a RoBERTa based sentiment classification model trained on a large corpus of 60 million English Tweets \cite{barbieri2020tweeteval} for initialization. We refer to this sentiment classification model as the TweetEval model for the rest of this paper. We use the Hugging Face implementation of the TweetEval sentiment classification model \footnote{\href{https://huggingface.co/cardiffnlp/twitter-roberta-base-sentiment}{https://huggingface.co/cardiffnlp/twitter-roberta-base-sentiment}}. The models are fine-tuned with a batch size of 16 and a learning rate of 2e-5. The TweetEval model pre-processes sentiment data to not include any URL's. We have done the same for for all the four datasets.

We compare our unsupervised model with a set of supervised models trained on each of the four datasets. We train supervised models by fine tuning the TweetEval model on each of the four datasets. Our experiments have shown that the TweetEval model performs the better in comparison to mBERT and XLM-RoBERTa based models for code-switched data.


\section{Evaluation}\label{section:eval}
We evaluate our results based on weighted average F1 and accuracy scores. When calculating the weighted average, the F1 scores are calculated for each class and a weighted average is taken based on the number of samples in each class. This metric is chosen because three out of four datasets we work with are highly imbalanced. We use the sklearn implementation for calculating weighted average F1 scores\footnote{ \url{https://scikit-learn.org/stable/modules/generated/sklearn.metrics.classification_report.html}}.

There are two ways to evaluate the performance of our proposed method, corresponding to two different perspectives with which we look at the outcome. One of the ways to evaluate the proposed method is to answer the question - `How good is the model when trained in the proposed, unsupervised manner?'. We call this perspective of evaluation, having a \textit{model perspective}. Here we're evaluating the strength of the unsupervised model. To evaluate the method from a model perspective, we compare the performance of best unsupervised model with the performance of the supervised model on the test set.

The second way to evaluate the proposed method is by looking at it from what we call an \textit{algorithmic perspective}. The aim of proposing an unsupervised algorithm is to be able to select sentences belonging to a particular sentiment class from an unkown dataset. Hence, to evaluate from an algorithmic perspective, we must look at the training set and check how accurate the algorithm is in its annotations for each class. To do this, we show performance (F1-scores) as a function of the number of selected sentences from the training set.


\begin{table}
\centering
\begin{tabular}{ccc}
\hline
\multicolumn{1}{|p{2cm}|}{\centering \textbf{Language}} & \multicolumn{1}{|p{1.5cm}|}{\centering \textbf{F1}} &  \multicolumn{1}{|p{1.5cm}|}{\centering \textbf{Accuracy}}\\
\hline
Hinglish & 0.32 & 0.36\\
Spanglish & 0.31 & 0.32\\
Tanglish &  0.15 & 0.16\\
Malayalam-English & 0.17 & 0.14\\
\hline
\end{tabular}
\caption{\label{Table:zeroshot}
Zero-shot prediction performance for the TweetEval model for each of the four datasets, for a two-class classification problem (positive and negative classes). The F1 scores represent the weighted average F1. These zero-shot predictions are for the training datasets in each of the four languages.
}
\end{table}

\begin{table*}
\centering
\begin{tabular}{|c||l|l||l|l||l|l||}
  \hline
  \multirow{2}{*}{Train Language} 
      & \multicolumn{2}{c||}{Vanilla} 
        & \multicolumn{2}{c||}{Ratio} 
                & \multicolumn{2}{c||}{Supervised} \\             \cline{2-7}
  & F1 & Accuracy & F1 & Accuracy & F1 & Accuracy\\  \hline
  $Hinglish$ & 0.84 & 0.84 & 0.84 & 0.84 & 0.91 & 0.91 \\      \hline
  $Spanglish$ & 0.77 & 0.76 & 0.77 & 0.77 & 0.78 & 0.79  \\      \hline
  $Tamil$ & 0.68 & 0.63 & 0.79 & 0.80 & 0.83 & 0.85  \\      \hline
  $Malayalam$ & 0.73 & 0.71 & 0.83 & 0.85 & 0.90 & 0.90\\      \hline
\end{tabular}
\caption{\label{Table:vanilla}
Performance of best Unsupervised Self-Training models for Vanilla and Ratio selection strategies when compared to performance of supervised models. The F1 scores represent the weighted average F1.
}
\end{table*}

\section{Experiments}\label{section:exp}
For our experiments, we restrict the dataset to consist of two sentiment classes - positive and negative sentiments. In this section, we evaluate our proposed unsupervised self-training framework for four different code-swithced languages spanning across three language families, with different dataset sizes and different extents of imbalance and code-mixing, and across two different domains. We also present a comparison between supervised and unsupervised models.

We present two training strategies for our proposed Unsupervised Self-Training Algorithm. The first is a vanilla strategy where the same number of sentences are selected in the selection block for each class. The second strategy uses a selection ratio - where we select sentences for fine tuning in a particular ratio from each class. We evaluate the algorithm based on the two evaluation criterion described in section \ref{section:eval}.

In the Fine-Tune Block in Figure \ref{fig:rec}, we fine-tune the TweetEval model on the selected sentences for only 1 epoch. We do this because we do not want our model to overfit on the small amount of selected sentences. This means that when we go through the entire training dataset, the model has seen every sentence in the train set \textit{exactly} once. We see that if we fine-tune the model for multiple epochs, the model overfits and its performance and capacity to learn reduces with every iteration.  

\subsection{Zero-Shot Results}
Table \ref{Table:zeroshot} shows the zero-shot results for the TweetEval model. We see that the zero-shot F1 scores are much higher for the Hinglish and Spanglish datasets when compared to the results for the Dravidian languages. Part of the disparity in zero-shot performance can be attributed to the differences in domains. This means that the TweetEval model is not as compatible to the Tanglish and Malayalam-English dataset than it is to the Spanglish and Hinglish datasets. Improved training strategies help increase performance. 

The zero-shot results in Table \ref{Table:zeroshot} use the TweetEval model, which is a 3-class classification model. Due to this, we get a prediction accuracy of less than 50\% for a binary classification problem. 

\subsection{Vanilla Selection}
In the Vanilla Selection strategy, we select the same number of sentences for each class. We saw no improvement when selecting less than 5\% sentences of the total dataset size in every iteration, equally split into the two classes. Table \ref{Table:vanilla} shows the performance of the best unsupervised model trained in comparison with a supervised model. For each of these results, N = 0.05 * (dataset size), where N/2 is the number of sentences selected from each class at every iteration step. The best unsupervised model is achieved almost halfway through the dataset for all languages. 

The unsupervised model performs surprisingly well for Spanglish when compared to the supervised counterpart. The performance for the Hinglish model is also comparable to the supervised model. This can be attributed to the fact that both datasets are in-domain for the TweetEval model and their zero-shot performances are better than for the Dravidian languages, as shown in Table \ref{Table:zeroshot}. Also, the fact that the Hinglish dataset is balanced helps improve performance. We expect the performance of the unsupervised models to increase with better training strategies.

For a balanced dataset like Hinglish, selecting N $>$ 5\% at every iteration provided similar performance whereas the performance deteriorates for the three imbalanced datasets if a larger number of sentences were selected. This behaviour was somewhat expected as when the dataset is imbalanced, the model is likely to make more errors in generating pseduo-labels for one class more than the other. Thus it helps to reduce the number of selections as that also reduces the number of errors. 


\subsection{Selection Ratio}
In this selection strategy, we select unequal number of samples from each class, deciding on a ratio of positive samples to negative samples. The aim of selecting sentences with a particular ratio is to incorporate the underlying class distribution of the dataset for selection. When the underlying distribution is biased, selecting equal number of sentences would leave the algorithm to have lower accuracy in the produced pseudo-labels for the smaller class, and this error is propagated with every iteration step.

The only way to truly estimate the correct selection ratio is to sample from the given dataset. In an unsupervised scenario, we would need to annotate a selected sample of sentences to determine the selection ratio empirically. We found that on sampling around 50 sentences from the dataset, we were accurately able to predict the distribution of the class labels with a standard deviation of approximately 4-6\%, depending on the dataset. The performance is not sensitive to estimation inaccuracies of that amount. 

Finding the selection ratio serves a second purpose - it also gives us an estimated stopping condition. By knowing an approximate ratio of the class labels and the size of the dataset, we now have an approximation for the total number of samples in each class. As soon as the total selections for a class across all iteration reaches the predicted number of samples of that class, according to the sampled selection ratio, we should stop the algorithm. 

The results for using the selection ratio are shown in Table \ref{Table:vanilla}. We see significant improvements in performance for the Dravidian languages, with the performance reaching very close to the supervised performance. The improvement in performance for the Hinglish and Spanglish datasets are minimal. This hints that a selection ratio strategy was able to overcome the difference in domains and the affects of poor initialization as pointed out in Table \ref{Table:zeroshot}.

\begin{figure}[ht]
  \centering
    \includegraphics[width=\linewidth]{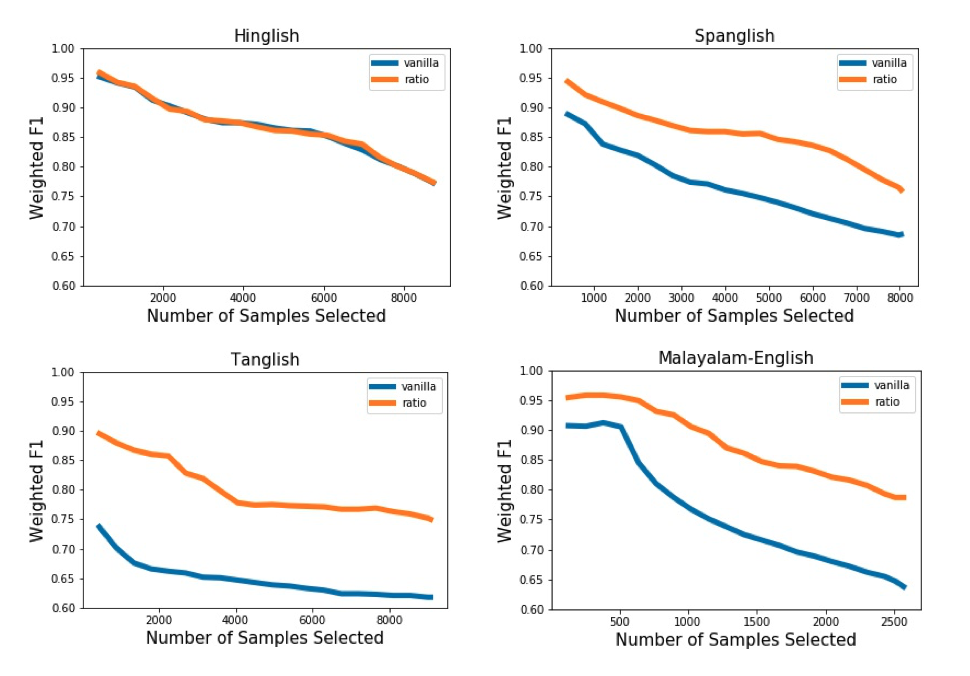}
    \caption{Performance of the Unsupervised Self-Training algorithm as a function of selected sentences from the training set.}\label{fig:algorithimic}
\end{figure}

The selection ratio strategy was also able to overcome the problem of data imbalance. This can be seen in Figure \ref{fig:algorithimic} when we evaluate the framework from an algorithmic perspective. Figure \ref{fig:algorithimic} plots the classification F1 scores of the unsupervised algorithm as a function of the selected sentences. We find that using the selection ratio strategy improves the performance on the training set significantly. We see improvements for the Dravidian languages, which were also reflected in Table \ref{Table:vanilla}. 

This improvement is also seen for the Spanglish dataset, which is not reflected in Table \ref{Table:vanilla}. This means that for Spanglish, the improvement in the unsupervised model when trained with selection ratio strategy does not generalize to a test set, but the improvement is enough to select sentences in the next iterations more accurately. This means that we're able to give better labels to our training set in an unsupervised manner, which was one of the aims of developing an unsupervised algorithm. (Note : The evaluation from a model perspective is done on the test set, whereas from an algorithmic perspective is done on the training set.)

This evaluation perspective also shows that if the aim of the unsupervised endevour is to create labels for an unlabelled set of sentences, one does not have to process the entire dataset. For example, if we are content with pseudo-labels or noisy labels for 4000 Hinglish Tweets, the accuracy of the produced labels would be close to 90\%.



\section{Analysis}\label{section:analysis}
In this section we aim to understand the information learnt by a model trained under the Unsupervised Self-Training. We take the example of the Hinglish dataset. To do so, we define a quantity called \textit{Token Ratio} to quantify the amount of code-mixing in a sentence. Since our initialization model is trained on an English dataset, the language of interest is Hindi and we would like to understand how well our model handles sentences with a large amount of Hindi. Hence, for the Hinglish dataset, we define the Hindi Token Ratio as:\\

\centerline{$\textit{Hindi Token Ratio} = \frac{ \textit{Number of Hindi Tokens} }{\textit{Total Number of Words} }$}

\begin{figure}[ht]
  \centering
    \includegraphics[width=0.99\linewidth]{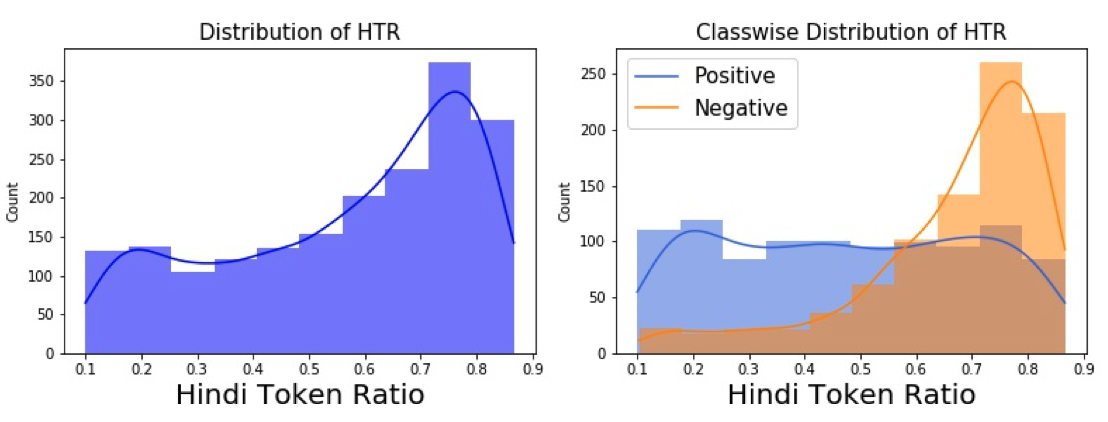}
    \caption{Distributions of Hindi Token Ratio for the Hinglish Dataset.}\label{fig:1}
\end{figure}

\cite{patwa2020semeval} provide three language labels for each token in the dataset - HIN, ENG, 0, where 0 usually corresponds to a symbol or other special characters in a Tweet. To quantify amount of code-mixing, we only use the tokens that have ENG or HIN as labels. Words are defined as tokens that have either the label HIN or ENG. We define the \textit{Token Ratio} quantity with respect to Hindi, but our analysis can be generalized to any code-mixed language. The distribution of the Hindi Token Ratio (HTR) in the Hinglish dataset is shown in Figure \ref{fig:1}. The figure clearly shows that the dataset is dominated by tweets that have a larger amount Hindi words than English words. This is also true for the other three datasets.

\subsection{Learning Dynamics of the Unsupervised Model}\label{sec:LD}
\begin{figure}[ht]
    \centering
    \includegraphics[width=\linewidth]{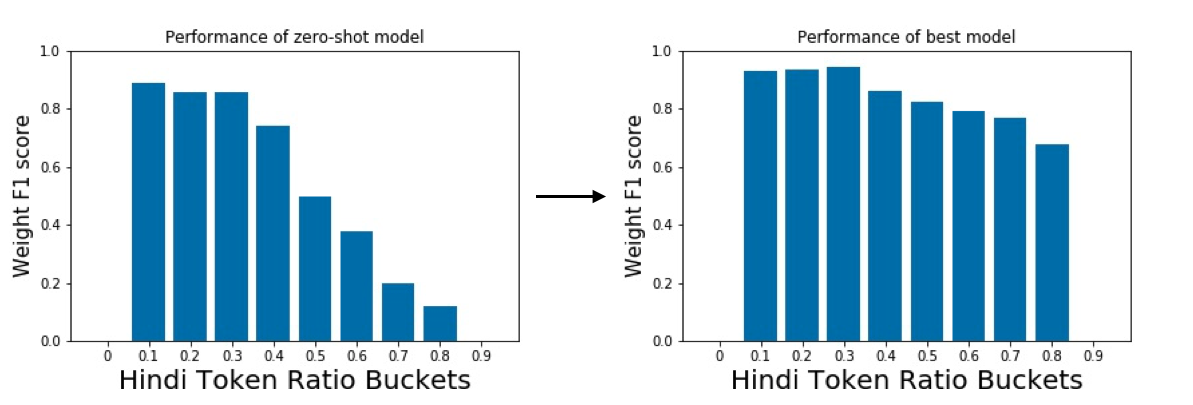}
    \caption{Model Performance for different Hindi Token Ratio buckets. For example, a bucket labelled as 0.3 contains all sentences that have a Hindi Token ration between 0.3 and o.4.}\label{fig:7}
\end{figure}

To study if the unsupervised model understands Hinglish, we look at the performance of the model as a function of the Hindi Token Ratio. In Figure \ref{fig:7} , the sentences in the Hinglish dataset are grouped into buckets of Hindi Token Ratio. A \textit{bucket} is of size 0.1 and contains all the sentences that fall in its range. For example, when the x-axis says 0.1, this means the bucket contains all sentences that have a Hindi Token ratio between 0.1 and 0.2.

Figure \ref{fig:7} shows that the zero shot model performs the best for sentences that have very low amount of Hindi code-mixed with English. As the amount of Hindi in a sentence increases, the performance of the zero-shot predictions decreases drastically. On training the model with our proposed Unsupervised Self-Training framework, we see a significant rise in the performance for sentences with higher HTR (or sentences with a larger amount of Hindi than English) as well as the overall performance of the model. This rise is gradual and the model improves at classifying sentences with higher HTR with every iteration.

\begin{figure}[ht]
    \centering
    \includegraphics[width=\linewidth]{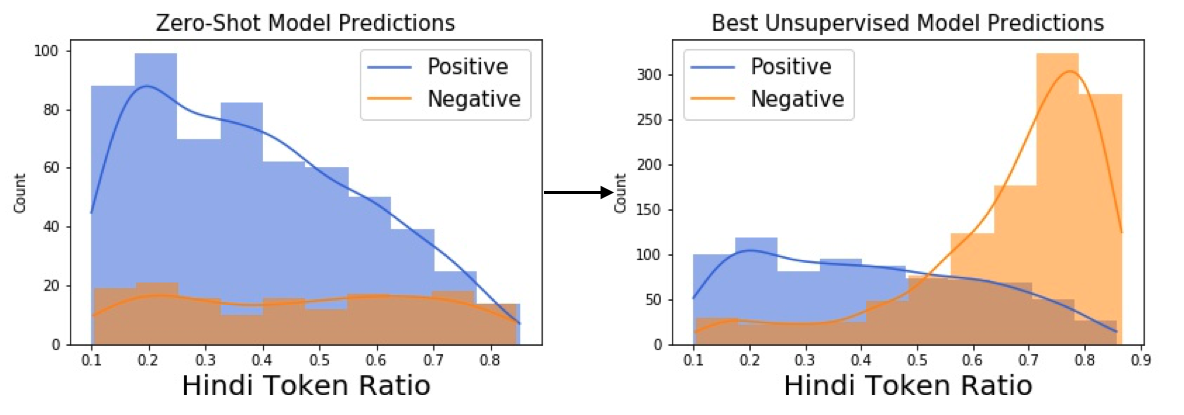}
    \caption{Comparison made between predictions made by the zero-shot and the best unsupervised model.}\label{fig:3}
\end{figure}

Next, we refer back to Figure \ref{fig:1}. Figure \ref{fig:1} shows the distribution of the Hindi Token Ratio for each of the two sentiment classes. For the Hinglish dataset, we see that tweets with negative sentiments are more likely to contain more Hindi words than English. The distribution for the Hindi Token Ratio for positive sentiment is almost uniform, thus showing no preference for English or Hindi words when expressing a positive sentiment. If we look at the distribution of the predictions made by the zero-shot unsupervised model, shown in Figure \ref{fig:3}, we see that majority of the sentences are predicted as belonging to the positive sentiment class. There seems to be no resemblance with the original distribution (Figure \ref{fig:1}). As the model trains under our Unsupervised Self-Training framework, we see that the predicted distribution becomes very similar to the original distribution.

\subsection{Error Analysis}
\begin{figure}[ht]
    \includegraphics[width=\linewidth]{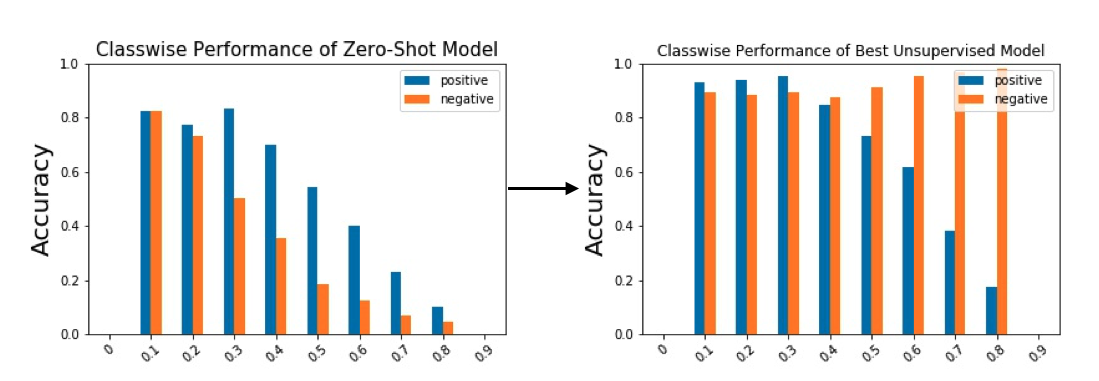}
    \caption{Class-wise performance of the zero-shot and the best unsupervised models for different Hindi Token Ratio buckets. }\label{fig:4}
\end{figure}

In this section, we look at the errors made by the unsupervised model. Table \ref{Table:error} shows the comparison between the class-wise performance of the supervised and the best unsupervised model. The unsupervised model is better at making correct predictions for the negative sentiment class when compared to the supervised model. Figure \ref{fig:4} shows the classwise performance for the zero-shot and best unsupervised model for the different HTR buckets. We see that the zero-shot models performs poorly for both the positive and negative classes. As the unsupervised model improves with iterations through the dataset, we see the performance for each class increase.
\begin{table}
\centering
\begin{tabular}{ccc}
\hline
\multicolumn{1}{|p{1cm}|}{\centering \textbf{Model Type}} & \multicolumn{1}{|p{2cm}|}{\centering \textbf{Positive Accuracy}} &  \multicolumn{1}{|p{2cm}|}{\centering \textbf{Negative Accuracy}}\\
\hline
Unsupervised & 0.73 & 0.94\\
Supervised & 0.93 & 0.83\\

\hline
\end{tabular}
\caption{\label{Table:error}
Comparison between class-wise performance for supervised and unsupervised models.
}
\end{table}



\subsection{Does the Unsupervised Model `Understand' Hinglish?}
The learning dynamics in section \ref{sec:LD} show that as the TweetEval model is fine-tuned under our proposed Unsupervised Training Framework, the performance of the model increases for sentences that have higher amounts of Hindi. In fact, the performance increase is seen across the board for all Hindi Token Ratio buckets. We saw the distribution of the predictions made by the zero-shot model, which preferred to classify almost all sentences as positive. But as the model was fine-tuned, the predicted distribution was able to replicate the original data distribution. These experiments show that the model originally trained on an English dataset is beginning to atleast recognize Hindi when trained with our proposed framework, if not understand it. 

We also see a bias in the Hinglish dataset where the negative sentiments are more likely to contain a larger number Hindi Tokens, which are unknown tokens from the perspective of the initial TweetEval model. Thus the classification task would be aided by learning the difference in the underlying distributions of the two classes. Note that we do expect a supervised model to use this divide in the distributions as well. Figure \ref{fig:4} shows a larger increase in performance for the negative sentiment class than the positive sentiment class, although the performance is increased across the board for all Hindi Token Ratio buckets. (This difference in performance can be remedied by selecting sentences with high Hindi Token Ratio in the selection block.) Thus, it does seem like that the model is able to understand Hindi and this understanding is aided by the differences in the class-wise distribution of the two sentiments.

\section{Conclusion}
We propose the Unsupervised Self-Training framework and show results for unsupervised sentiment classification of code-switched data. The algorithm is comprehensively tested for four very different code-mixed languages - Hinglish, Spanglish, Tanglish and Malayalam-English, covering many variations including differences in language families, domains, dataset sizes and dataset imbalances. The unsupervised models performed competitively when compared to supervised models. We also present training strategies to optimize the performance of our proposed framework. 

An extensive analysis is provided describing the learning dynamics of the algorithm. The algorithm is initialized with a model trained on an English dataset and has poor zero-shot performance on sentence with large amounts of code-mixing. We show that with every iteration, the performance on fine-tuned model increases for sentences with a larger amounts of code-mixing. Eventually, the model begins to understand the code-mixed data.

\section{Future Work}
The proposed Unsupervised Self-Training algorithm was tested with only two sentiment classes - positive and negative. An unsupervised sentiment classification algorithm is to be able to generate annotations for an unlabelled code-mixed dataset without going through the expensive annotation process. This can be done by including the neutral class in the dataset, which is going to be a part of our future work. 


In our work, we only used one initialization model trained on English Tweets for all four code-mixed datasets, as all of them were code-mixed with English. Future work can include testing the framework with different and more compatible models for initialization. Further work can be done on optimization strategies, including incorporating the Token Ratio while selecting pseudo-labels, and active learning.



\bibliography{anthology,custom}
\bibliographystyle{acl_natbib}

\newpage
\appendix
\section{Implementation Details}
We use the RoBERTa-base based model pre-trained on a large English Twitter corpus for initialization, which has about 125M paramters. The model was fine-tuned using the NVIDIA GeForce GTX 1070 GPU using python3.6. The Tanglish dataset was the biggest dataset which required approximately 3 minutes per iteration. One pass through the entire dataset required 20 iterations for the Vanilla selection strategy and about 30 iterations for the Ratio selection strategy. The time required per iteration was lower for the the other three datasets, with about 100 seconds per iteration for the Malaylam-English datasets.

\end{document}